\documentclass[letterpaper]{article} 
\usepackage{aaai2026}  
\usepackage{times}  
\usepackage{helvet}  
\usepackage{courier}  
\usepackage[hyphens]{url}  
\usepackage{graphicx} 
\usepackage{natbib}  
\usepackage{caption} 
\usepackage{algorithm}
\usepackage{algorithmic}

\usepackage{newfloat}
\usepackage{listings}
\DeclareCaptionStyle{ruled}{labelfont=normalfont,labelsep=colon,strut=off} 
\floatstyle{ruled}
\newfloat{listing}{tb}{lst}{}
\floatname{listing}{Listing}
\usepackage{amssymb}
\usepackage{amsmath}
\usepackage{multirow}
\usepackage{array}
\usepackage{booktabs}
\usepackage{subcaption}
\usepackage{xcolor}

\newcolumntype{C}[1]{>{\centering\arraybackslash}m{#1}}
\newcolumntype{N}{>{\centering\arraybackslash}p{0.9cm}}
\newcolumntype{L}[1]{>{\raggedright\arraybackslash}m{#1}}

\title{Efficient Hallucination Detection: Adaptive Bayesian Estimation of Semantic Entropy with Guided Semantic Exploration}
\author{
    Qiyao Sun\textsuperscript{\rm 1},
    Xingming Li\textsuperscript{\rm 1},
    Xixiang He\textsuperscript{\rm 1},
    Ao Cheng\textsuperscript{\rm 1},
    Xuanyu Ji\textsuperscript{\rm 1},\\
    Hailun Lu\textsuperscript{\rm 2},
    Runke Huang\textsuperscript{\rm 3},
    Qingyong Hu\textsuperscript{\rm 2}\thanks{Corresponding author.}
}
\affiliations{
    \textsuperscript{\rm 1}National University of Defense Technology, Changsha, China\\
    \textsuperscript{\rm 2}Intelligent Game and Decision Lab, Beijing, China\\
    \textsuperscript{\rm 3}The Chinese University of Hong Kong, Shenzhen, China\\

    \{sunqiyao18, lixingming, hexixiang, chengao18, jixuanyu18\}@nudt.edu.cn, Luhailun0728@outlook.com, runkehuang@cuhk.edu.cn, huqingyong15@outlook.com
}

\usepackage{bibentry}

\begin{document}

\maketitle

\begin{abstract}
Large language models (LLMs) have achieved remarkable success in various natural language processing tasks, yet they remain prone to generating factually incorrect outputs—known as ``hallucinations". While recent approaches have shown promise for hallucination detection by repeatedly sampling from LLMs and quantifying the semantic inconsistency among the generated responses, they rely on fixed sampling budgets that fail to adapt to query complexity, resulting in computational inefficiency. We propose an Adaptive Bayesian Estimation framework for Semantic Entropy with Guided Semantic Exploration, which dynamically adjusts sampling requirements based on observed uncertainty. Our approach employs a hierarchical Bayesian framework to model the semantic distribution, enabling dynamic control of sampling iterations through variance-based thresholds that terminate generation once sufficient certainty is achieved. We also develop a perturbation-based importance sampling strategy to systematically explore the semantic space. Extensive experiments on four QA datasets demonstrate that our method achieves superior hallucination detection performance with significant efficiency gains. In low-budget scenarios, our approach requires about 50\% fewer samples to achieve comparable detection performance to existing methods, while delivers an average AUROC improvement of 12.6\% under the same sampling budget.
\end{abstract}


\section{Introduction}

Large language models (LLMs) \cite{arxiv2025surveylargelanguagemodels} have demonstrated remarkable capabilities in language understanding, generation, and reasoning, fundamentally transforming the landscape of natural language processing \cite{emnlp2023beyondfactuality, acmcomputsurv2024surveyonautomaticgeneration}. However, these models exhibit a critical limitation: they are prone to generating hallucination contents that appear plausible and coherent but lack factual grounding or contradict verifiable information \cite{corr2023surveyfactualityinllm}. Unlike traditional natural language generation systems where hallucinations primarily involve inconsistencies with source content, LLM hallucinations encompass a broader spectrum of factual errors and faithfulness issues due to their open-ended nature \cite{acmtrans2025surveyhallucination}. The human-like fluency of LLM responses makes these hallucinations particularly difficult to detect, creating significant risks when deploying these models in real-world scenarios such as education, economics, science and so on \cite{corr2024surveyllminfinance, corr2024surveyllm4edu, corr2025surveyllm4sci}. This fundamental challenge poses a substantial barrier to the reliable integration of LLMs into critical information systems where accuracy and trustworthiness are paramount.

\begin{figure}[t]
\centering
\includegraphics[width=.90\columnwidth]{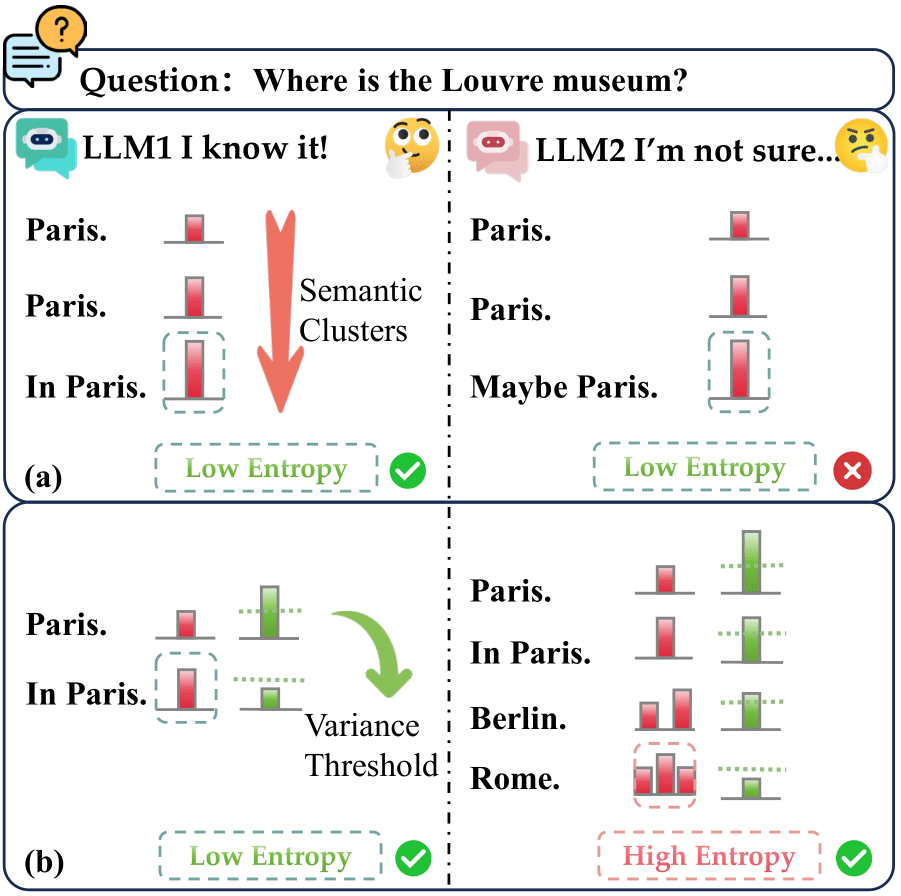}
\caption{Comparison of fixed sampling (a) versus our adaptive Bayesian approach (b) for hallucination detection. Fixed sampling wastes computational resources on simple queries (LLM1) while failing to discover semantic diversity in complex cases (LLM2). Our method dynamically adjusts sampling based on variance thresholds, enabling efficient and accurate hallucination detection.}
\end{figure}

Current hallucination detection methods can be broadly categorized into four paradigms based on their underlying mechanisms. (1) External knowledge-based methods leverage retrieval systems to validate LLM outputs against authoritative sources \cite{emnlp2023kcts,  aclfindings2024chainofverification}, but face limitations in domain coverage and require complex verification pipelines. (2) Metacognition-based methods prompt LLMs to assess their own confidence \cite{corr2022ptrue}, yet struggle with unreliability. (3) Single-sample methods analyze token-level patterns or hidden states within individual generations \cite{corr2024semanticentropyprobes, emnlp2023focus}, offering computational efficiency but limited accuracy. (4) Multi-sample methods, particularly semantic entropy approaches \cite{nature2024semanticentropy}, estimate uncertainty by clustering semantically equivalent outputs across multiple samples, demonstrating superior detection performance compared to other methods.

Despite their effectiveness, existing multi-sample methods suffer from a critical limitation: they employ fixed sampling budgets that fail to adapt to the inherent complexity of different queries. Simple factual questions may require only a few samples to reliably estimate semantic uncertainty, while complex or ambiguous queries necessitate extensive exploration of the semantic space. This one-size-fits-all approach results in computational waste for straightforward queries and insufficient sampling for challenging ones. Furthermore, current methods rely on multinomial sampling, which may repeatedly generate semantically similar outputs without efficiently exploring the full distribution of possible meanings. The lack of adaptive mechanisms and guided exploration strategies fundamentally limits the practical deployment of these methods in low-budget settings.

To address these limitations, we propose an \textbf{Adaptive Bayesian Estimation framework} for Semantic Entropy with \textbf{Guided Semantic Exploration}. Our approach introduces a hierarchical Bayesian framework that explicitly models the semantic distribution through a Dirichlet prior and decomposes the semantic entropy expectation via marginalization into two components: the posterior over the number of semantic categories and the conditional entropy given each possible category count. Building on this probabilistic foundation, we implement variance-based adaptive sampling that dynamically adjusts the number of queries based on posterior uncertainty, enabling efficient allocation of computational resources according to problem complexity. To accelerate variance convergence and enhance sampling efficiency, we develop a guided exploration strategy that identifies semantically critical tokens through importance weighting and systematically perturbs them to discover diverse interpretations, while employing importance sampling to maintain unbiased estimates.

Extensive experiments on four QA datasets demonstrate that our method achieves comparable or superior hallucination detection performance while significantly reducing the required number of model queries.

The main contributions can be summarized as follows:

\begin{itemize}
    \item We introduce an adaptive Bayesian framework for semantic entropy estimation that dynamically adjusts sampling requirements based on observed uncertainty.
    \item We develop a guided semantic exploration strategy with importance sampling that discovers diverse semantic interpretations by perturbing critical tokens, accelerating convergence compared to random sampling.
    \item We demonstrate through comprehensive experiments that our method achieves state-of-the-art hallucination detection performance with significantly fewer samples, particularly excelling in resource-constrained scenarios.
\end{itemize}

\section{Related Work}

\subsection{Hallucination}

Hallucination in large language models refers to the generation of content that appears coherent but lacks factual support or deviates from the intended output \cite{corr2023surveyfactualityinllm}. Current research categorizes LLM hallucinations into two main types: factuality hallucination and faithfulness hallucination \cite{acmtrans2025surveyhallucination}. Factuality hallucination occurs when generated content contains errors that contradict verifiable facts, while faithfulness hallucination manifests as inconsistencies with provided context or internal logic within the generated text. The ability of LLMs to produce highly convincing and human-like responses makes detecting these hallucinations particularly challenging, necessitating the development of robust detection methods.

\subsection{Hallucination Detection}

The detection of hallucinations in LLMs has emerged as a critical research area, with various approaches developed to identify when models generate factually incorrect or inconsistent content. Current detection methods can be broadly categorized into three main paradigms based on their underlying mechanisms.

\paragraph{External knowledge based methods} use external knowledge bases to guide LLM generation and validate outputs \cite{emnlp2023kcts, aclfindings2024chainofverification, emnlp2023factscore, emnlp2023haludetbybayesiansequentialestimation}. However, these approaches face significant challenges as they heavily depend on the accuracy and completeness of external knowledge sources, and struggle with domain-specific or rapidly evolving information.

\paragraph{Metacognition based methods} utilize the metacognitive capabilities of LLMs—their ability to reflect on and evaluate their own knowledge states. The fundamental assumption is that LLMs develop implicit metacognitive signals during training that enable them to distinguish between confident, factual generations and uncertain, potentially hallucinated content \cite{corr2022ptrue}. These methods face challenges in calibrating self-awareness across diverse domains and struggle when the model's metacognitive judgments are themselves unreliable.

\paragraph{Single-sample based methods} require only a single forward pass through the model. FOCUS \cite{emnlp2023focus} retrieve relevant tokens from the generation process and analyze token-level confidence patterns. While \cite{corr2024semanticentropyprobes} train specialized classifiers to predict uncertainty metrics directly from the model's hidden states. These methods offer computational efficiency but often face challenges in interpretability and accuracy.

\paragraph{Multi-sample based methods} detect hallucinations by analyzing inconsistencies across multiple LLM outputs. Early approaches using Lexical Similarity \cite{acl2022lexsim} and Predictive Entropy \cite{corr2022ptrue} often overestimated uncertainty due to diverse surface forms. Semantic Entropy \cite{iclr2023semanticentropy, nature2024semanticentropy} marked a breakthrough by clustering outputs into semantic equivalence classes and computing entropy over meaning distributions. While subsequent refinements \cite{acl2024sar, acl2024mars, iclr2024eigenscore, nips2024kle} and SDLG's \cite{iclr2025sdlg} diversified exploration strategies improved performance, all existing methods rely on fixed sampling budgets that ignore query complexity. Our adaptive framework addresses this limitation by dynamically adjusting sample sizes based on inherent uncertainty, achieving computational efficiency with detection accuracy improvement.

\section{Problem Formulation}

\paragraph{Language Model Generation} Let $\mathcal{X}$ denote the space of all possible prompts that can be presented to a language model. For a given prompt $x \in \mathcal{X}$, let $\mathcal{R}_x$ represent the set of all possible response sequences that the LLM can generate. We model the LLM's generation process as a conditional probability distribution $P_\theta(\mathbf{r}|x)$, where $\theta$ represents the model parameters and $\mathbf{r} \in \mathcal{R}_x$ is a response sequence.

\paragraph{Semantic Equivalence} We define $\mathcal{M}_x$ as the set of distinct semantic meanings for prompt $x$. Let $f_x: \mathcal{R}_x \rightarrow \mathcal{M}_x$ be a mapping function that assigns each response $r \in \mathcal{R}_x$ to its corresponding meaning class $m \in \mathcal{M}_x$. This mapping induces a partition over the response space, where responses within the same partition are semantically equivalent.

\paragraph{Semantic Entropy} Given a prompt $x$ and the LLM's response distribution $P_\theta(\mathbf{r}|x)$, The probability of generating meaning $m$ is:

\begin{equation}
    p(m|x) = \sum_{r \in \mathcal{R}_x: f_x(r) = m} P_\theta(r|x)
\end{equation}

The semantic entropy for prompt $x$ is then defined as the Shannon entropy over the meaning distribution:

\begin{equation}
    H_{sem} = -\sum_{m \in \mathcal{M}_x} p(m|x) \log p(m|x)
\end{equation}

\paragraph{Estimation Problem} In practice, we cannot enumerate all possible responses in $\mathcal{R}_x$ to compute the exact semantic entropy. Instead, treating $H_{sem}$ as a random variable, we estimate it from a finite dataset obtained by sampling from the LLM. For a given prompt $x$, we generate $N$ independent response sequences $r_1, \ldots, r_N \sim P_\theta(\cdot|x)$. 

For each sampled sequence $r_i$, we determine its semantic meaning $m_i = f_x(r_i)$, yielding a corresponding list of meanings $m_1, \ldots, m_N$. Additionally, we obtain the generation probability $P_\theta(r_i|x)$ for each sequence. Thus, the estimation dataset is defined as:
\begin{equation}
    \mathcal{D} = \{(r_1, m_1, P_\theta(r_1|x)), \ldots, (r_N, m_N, P_\theta(r_N|x))\}
\end{equation}

We seek to develop an efficient semantic entropy estimator that dynamically increases $N$ based on the observed sample variance, ensuring reliable yet minimal sampling cost.

\begin{algorithm}[tb]
\caption{Bayesian Estimation of Semantic Entropy}
\label{alg:adaptive_semantic_entropy}
\textbf{Input}: Prompt $x$, LLM $P_\theta$, variance threshold $\gamma$\\
\textbf{Parameter}: Initial samples $N_0$, top-$k$ alternatives\\
\textbf{Output}: Semantic entropy $\hat{H}_{sem}$
\begin{algorithmic}[1]
\STATE \textbf{// Initialize with weighted perplexity prior}
\STATE Generate initial samples $\{r_1, \ldots, r_{N_0}\} \sim P_\theta(\cdot|x)$
\STATE Compute token importance weights $w_{i,j}$ \hfill $\boldsymbol{\triangleright}$ \textbf{Eq. (15)}
\STATE Calculate weighted perplexity $\hat{\lambda}$ \hfill $\boldsymbol{\triangleright}$ \textbf{Eq. (17)}
\STATE Set prior $p(K) \sim \text{Poisson}(\hat{\lambda})$
\STATE \textbf{// Initialize dataset with initial samples}
\STATE $\mathcal{D} \leftarrow \{(r_i, m_i, P_\theta(r_i|x)) : i = 1, \ldots, N_0\}$
\STATE Compute initial $\mathbb{E}[\mathbf{h}|\mathcal{D}]$ and $\text{Var}[\mathbf{h}|\mathcal{D}]$ \hfill $\boldsymbol{\triangleright}$ \textbf{Eq. (4-5)}
\STATE \textbf{// Adaptive sampling loop}
\WHILE{$\text{Var}[\mathbf{h}|\mathcal{D}] > \gamma$}
    \STATE Generate new sample $(r', w')$ via \textbf{Guided Semantic Exploration} from random $r \in \mathcal{D}$ using top-$k$\\ \hfill $\boldsymbol{\triangleright}$ \textbf{Section 4.2}
    \STATE Add $(r', m', P_\theta(r'|x))$ to $\mathcal{D}$
    \STATE \textbf{// Bayesian update}
    \STATE Update effective counts $n_j$ and Dirichlet posterior parameters $\tilde{\alpha}_j$ with importance weight $w'$\\ \hfill $\boldsymbol{\triangleright}$ \textbf{Eq. (23-24)}
    \STATE Update truncated Dirichlet posterior constraints $\mathcal{C}$\\ \hfill $\boldsymbol{\triangleright}$ \textbf{Eq. (6-7)}
    \STATE Compute posterior $p(K|\mathcal{D})$ \hfill $\boldsymbol{\triangleright}$ \textbf{Eq. (11)}
    \STATE Recalculate $\mathbb{E}[\mathbf{h}|\mathcal{D}]$ and $\text{Var}[\mathbf{h}|\mathcal{D}]$ \hfill $\boldsymbol{\triangleright}$ \textbf{Eq. (4-5)}
\ENDWHILE
\STATE \textbf{return} $\hat{H}_{sem} = \mathbb{E}[\mathbf{h}|\mathcal{D}]$
\end{algorithmic}
\end{algorithm}

\section{Method}

We propose a hierarchical Bayesian framework for efficient semantic entropy estimation that addresses two fundamental challenges: uncertainty in the number of distinct semantic meanings and the computational cost of exhaustive sampling. Our approach decomposes the estimation problem by marginalizing over the unknown cardinality of the semantic space, employing a truncated Dirichlet posterior that incorporates generation probability constraints to provide tighter uncertainty bounds. To adaptively calibrate prior beliefs, we introduce a weighted perplexity metric that captures prompt-specific semantic diversity. Furthermore, we develop a guided semantic exploration strategy using importance sampling, which systematically perturbs semantically critical tokens to discover diverse interpretations while maintaining unbiased estimates. The framework dynamically adjusts sampling based on observed variance, enabling reliable semantic entropy estimation with minimal cost. We detail the hierarchical Bayesian framework (Section 4.1) and the guided exploration mechanism (Section 4.2) below.

\subsection{Hierarchical Bayesian Framework}

We propose a hierarchical Bayesian framework that stratifies the semantic entropy estimation based on the number of distinct semantic categories $|\mathcal{M}_x|$. Given the dataset $\mathcal{D}$ containing sampled responses and their semantic meanings, the key insight is that the semantic entropy $H_{sem}$ fundamentally depends on both the cardinality of the meaning space and the probability distribution over these meanings.

Let $K = |\mathcal{M}_x|$ denote the number of distinct semantic meanings in $\mathcal{M}_x$, and let $\mathbf{h}$ be a random variable that represents our belief about the value of $H_{sem}$. To compute its expected value given the observed dataset $\mathcal{D}$, we apply the law of total expectation to marginalize over the unknown number of semantic categories:

\begin{equation}
    \mathbb{E}[\mathbf{h}|\mathcal{D}] = \sum_{K=1}^{\infty} \mathbb{E}[\mathbf{h}|K,\mathcal{D}] \cdot p(K|\mathcal{D})
\end{equation}

\begin{equation}
    \text{Var}[\mathbf{h}|\mathcal{D}] = \mathbb{E}_K[\text{Var}[\mathbf{h}|K,\mathcal{D}]] + \text{Var}_K[\mathbb{E}[\mathbf{h}|K,\mathcal{D}]]
\end{equation}

This hierarchical decomposition naturally separates the estimation problem into two components. $\mathbb{E}[\mathbf{h}|K,\mathcal{D}]$ represents the expected entropy conditioned on exactly $K$ distinct semantic meanings being present in $\mathcal{M}_x$. This expectation is taken with respect to the posterior distribution of $\mathbf{p}$ given both $K$ and the observed data. $p(K|\mathcal{D})$ captures our posterior belief about the cardinality of the meaning space after observing the sampled responses.

To enable adaptive sampling, we employ the total variance $\text{Var}[\mathbf{h}|\mathcal{D}]$ from Equation (5) as our stopping criterion. The sampling process terminates when $\text{Var}[\mathbf{h}|\mathcal{D}] < \gamma$, where $\gamma$ is a predefined threshold that controls the trade-off between estimation accuracy and computational cost.

\subsubsection{Calculation of $\mathbb{E}[\mathbf{h}|K,\mathcal{D}]$ and $\text{Var}[\mathbf{h}|K,\mathcal{D}]$}

Given a fixed number of semantic categories $K$, we model the probability distribution over these categories using a Dirichlet prior. Let $\mathbf{p} = (p_1, \ldots, p_K)$ denote the probability vector where $p_j$ represents the probability of generating semantic meaning $j \in \{1, \ldots, K\}$. We adopt a uninformative Dirichlet prior $\mathbf{p} \sim \text{Dir}(\alpha_0, \ldots, \alpha_0)$.

After observing the dataset $\mathcal{D}$, let $n_j = |\{i : m_i = j, i = 1, \ldots, N\}|$ denotes the number of sampled responses that map to semantic meaning $j$. Under standard Bayesian updating, the posterior would be $\text{Dir}(\alpha_0 + n_1, \ldots, \alpha_0 + n_K)$.

However, the LLM's generation probabilities provide additional constraints on the feasible probability space. For each semantic category $j$, the true probability $p_j$ satisfy:

\begin{equation}
    p_j \geq \sum_{\substack{r_i \in \mathcal{D}:\: f_x(r_i) = j}} P_\theta(r_i|x) \triangleq b_j
\end{equation}

This constraint arises because we have directly observed specific sequences belonging to category $j$ with their generation probabilities. The constraint set is thus defined as:

\begin{equation}
    \mathcal{C} = \{\mathbf{p} \in \Delta^{K-1} : p_j \geq b_j \text{ for all } j = 1, \ldots, K\}
\end{equation}

where $\Delta^{K-1}$ denotes the $K-1$ dimensional probability simplex.

The posterior distribution becomes a truncated Dirichlet distribution. Let $\pi(\mathbf{p})$ denote the density of $\text{Dir}(\boldsymbol{\alpha})$ where $\boldsymbol{\alpha} = (\alpha_0 + n_1, \ldots, \alpha_0 + n_K)$. The truncated Dirichlet distribution over $\mathcal{C}$ has density:

\begin{equation}
    \pi_{\mathcal{C}}(\mathbf{p}) = \begin{cases}
        \frac{\pi(\mathbf{p})}{Z_{\mathcal{C}}} & \text{if } \mathbf{p} \in \mathcal{C} \\
        0 & \text{otherwise}
    \end{cases}
\end{equation}

where $Z_{\mathcal{C}} = \int_{\mathcal{C}} \pi(\mathbf{p}) d\mathbf{p}$ is the normalization constant.

The expected semantic entropy given $K$ and $\mathcal{D}$ is:

\begin{equation}
    \mathbb{E}[\mathbf{h}|K,\mathcal{D}] = \int H(\mathbf{p}) \cdot \pi_{\mathcal{C}}(\mathbf{p}) d\mathbf{p}
\end{equation}

where $H(\mathbf{p}) = -\sum_{j=1}^K p_j \log p_j$ is the Shannon entropy. Similarly, the variance is:

\begin{equation}
    \text{Var}[\mathbf{h}|K,\mathcal{D}] = \int H^2(\mathbf{p}) \cdot \pi_{\mathcal{C}}(\mathbf{p}) d\mathbf{p} - \mathbb{E}^2[\mathbf{h}|K,\mathcal{D}]
\end{equation}

In practice, these integrals are computed via self-normalized importance sampling \cite{nips2015snis}. The detailed implementation is discussed in Appendix A.

\subsubsection{Posterior Inference of $p(K|\mathcal{D})$}

To compute the posterior distribution $p(K|\mathcal{D})$, we apply Bayes' theorem:
\begin{equation}
    p(K|\mathcal{D}) = \frac{p(\mathcal{D}|K) \cdot p(K)}{\sum_{K'=1}^{\infty} p(\mathcal{D}|K') \cdot p(K')}
\end{equation}

\textbf{Calculation of marginal likelihood $p(\mathcal{D}|K)$} Now we compute the marginal likelihood $p(\mathcal{D}|K)$, which represents the probability of observing the dataset $\mathcal{D}$ given $K$ semantic categories. This requires marginalizing over all possible probability distributions $\mathbf{p}$ consistent with $K$ categories:
\begin{equation}
    p(\mathcal{D}|K) = \int p(\mathcal{D}|\mathbf{p}, K) \cdot \pi_{\mathcal{C}}(\mathbf{p}) d\mathbf{p}
\end{equation}

Given the observed category counts $\mathbf{n} = (n_1, \ldots, n_K)$ from the sampled responses, the likelihood follows a multinomial distribution:
\begin{equation}
    p(\mathcal{D}|\mathbf{p}, K) = \frac{N!}{\prod_{j=1}^K n_j!} \prod_{j=1}^K p_j^{n_j}
\end{equation}

The integral in Equation (12) can be efficiently evaluated using importance sampling employed for Equation (9).

\textbf{Calculation of prior $p(K)$ via weighted perplexity} We model the unknown number of semantic classes $K$ using a Poisson prior with parameter $\lambda$:
\begin{equation}
    p(K) = \frac{\lambda^K e^{-\lambda}}{K!}
\end{equation}

This choice reflects the view that distinct semantic meanings emerge as rare events in the LLM's vast semantic space. The parameter $\lambda$ controls the expected number of coherent semantic interpretations for a given prompt.

we propose an adaptive elicitation method that calibrates $\lambda$ based on the LLM's inherent uncertainty for the specific prompt, which begins by generating a small initial sample of $N_0$ responses $\{r_1, \ldots, r_{N_0}\}$ from $P_\theta(\cdot|x)$. For each response $r_i$, we quantify the semantic importance of individual tokens to capture their contribution to the overall meaning. Let $r_i = (t_{i,1}, \ldots, t_{i,L_i})$ denote the token sequence of length $L_i$. We compute the importance weight for token $t_{i,j}$ as:

\begin{equation}
    w_{i,j} = 1 - \text{sim}(r_i, r_i \setminus \{t_{i,j}\})
\end{equation}

where $\text{sim}$ measures the similarity between two sentences on a scale of 0 to 1 and $r^{(0)}_i \setminus \{t_{i,j}\}$ denotes the response with token $t_{i,j}$ removed.

Using these importance weights, we compute a weighted perplexity for each response that emphasizes semantically critical tokens:

\begin{equation}
    \text{WPL}_i = \exp\left(-\frac{\sum_{j=1}^{L_i} w_{i,j} \log P_\theta(t_{i,j}|t_{i,<j}, x)}{\sum_{j=1}^{L_i} w_{i,j}}\right)
\end{equation}

where $P_\theta(t_{i,j}|t_{i,<j}, x)$ is the conditional probability of token $t_{i,j}$ given the preceding tokens and prompt.

The empirical estimate of $\lambda$ is then obtained as:

\begin{equation}
    \hat{\lambda} = \frac{1}{N_0} \sum_{i=1}^{N_0} \text{WPL}_i
\end{equation}

This weighted perplexity serves as a principled proxy for semantic diversity: higher values indicate greater uncertainty in the model's semantic space, suggesting more potential semantic categories.

To handle the infinite summation in Equation (11), we employ a truncation strategy. Since $p(K)$ decays exponentially for large $K$ under the Poisson prior, we truncate the summation at $K_{max} = \max(K_{obs}, 3\lambda)$, where $K_{obs}$ denotes the number of distinct semantic meanings observed in $\mathcal{D}$.

\subsection{Guided Semantic Exploration}

To improve the efficiency of semantic entropy estimation, we develop a guided exploration strategy that leverages importance sampling to systematically explore the LLM's semantic space. By strategically perturbing tokens at semantically critical positions and continuing generation from these points, we can discover diverse semantic interpretations while maintaining computational efficiency.

\subsubsection{Guided Language Generation}

We employ a perturbation-based approach to construct sequences that explore alternative semantic branches. For a given response $r = (t_1, \ldots, t_L) \sim P_\theta(\cdot|x)$, we first identify semantically critical positions using the token importance weights defined in Equation (15). Let $\mathcal{I} = \{i_1, i_2, \ldots, i_L\}$ denote the indices of all tokens ordered by their importance weights in descending order, where $w_{i_1} \geq w_{i_2} \geq \ldots \geq w_{i_L}$.

At each critical position $i_j \in \mathcal{I}$, we examine the conditional token distribution $P_\theta(\cdot|t_{<i_j}, x)$ and identify the top-$k$ alternative tokens excluding the original choice:
\begin{equation}
    \mathcal{A}_{i_j} = \text{top-}k\{t \in \mathcal{V} \setminus \{t_{i_j}\} : P_\theta(t|t_{<i_j}, x)\}
\end{equation}

where $\mathcal{V}$ denotes the vocabulary. For each alternative token $t' \in \mathcal{A}_{i_j}$, we generate a new response by replacing $t_{i_j}$ with $t'$ and continuing the generation:
\begin{equation}
    r' = (t_1, \ldots, t_{i_j-1}, t', t'_{i_j+1}, \ldots, t'_{L'}) 
\end{equation}

where $t'_{i_j+1}, \ldots, t'_{L'}$ are sampled from $P_\theta(\cdot|t_1, \ldots, t_{i_j-1}, t', x)$.

\subsubsection{Importance Sampling}

The guided semantic exploration process described above forcibly modifies certain tokens in the generated sequences, deviating from the LLM's original distribution $P_\theta(\cdot|x)$. Since we are no longer sampling directly from $P_\theta(\cdot|x)$, we must correct for this bias through importance sampling with a properly defined proposal distribution $q(\mathbf{r}|x)$ that accurately captures our modified sampling procedure.

The proposal distribution is defined as:
\begin{equation}
    q(\mathbf{r}|x) = \sum_{r' \in \mathcal{R}_x} p(\mathbf{r}|r', x) \cdot P_\theta(r'|x)
\end{equation}

where $p(\mathbf{r}|r', x)$ represents the probability of transforming an initial response $r'$ into $\mathbf{r}$ through our guided generation process. Under specific assumptions about token selection \cite{iclr2025sdlg}, the proposal distribution takes the form:
\begin{equation}
    q(\mathbf{r}|x) = \frac{P_\theta(\mathbf{r}|x)}{P_\theta(t_j|\mathbf{t}_{<j}, x)}
\end{equation}

where $j$ is the index of the perturbed token in sequence $\mathbf{r} = (t_1, \ldots, t_L)$.

The importance weight for a sample $\mathbf{r}$ drawn from $q(\cdot|x)$ is:
\begin{equation}
    w(\mathbf{r}) = \frac{P_\theta(\mathbf{r}|x)}{q(\mathbf{r}|x)} = P_\theta(t_j|\mathbf{t}_{<j}, x)
\end{equation}

This weight ensures unbiased estimation while promoting exploration of lower-probability but potentially semantically distinct sequences (See Appendix B for proof).

\subsubsection{Bayesian Update with Weighted Samples}

When incorporating samples obtained through importance sampling into our hierarchical Bayesian framework, we must properly account for their importance weights. Let $\{(r^{(1)}, w^{(1)}), \ldots, (r^{(N-1)}, w^{(N-1)})\}$ denote the sequence of weighted samples, where $r^{(t)}$ is drawn from $q(\cdot|x)$ with importance weight $w^{(t)}$.

We modify the effective counts to reflect the importance weights. If sample $r^{(N)}$ is assigned to semantic category $j$, the effective count update becomes:
\begin{equation}
    n_j^{(N)} = n_j^{(N-1)} + w^{(N)}
\end{equation}

The Dirichlet posterior parameters are then updated as:
\begin{equation}
    \alpha_j^{(N)} = \alpha_0 + n_j^{(N)}
\end{equation}

To maintain proper normalization, we scale the posterior parameters:
\begin{equation}
    \tilde{\alpha}_j^{(N)} = \alpha_j^{(N)} \cdot \frac{\sum_{k=1}^{K} \alpha_k^{(0)} + N}{\sum_{k=1}^{K} \alpha_k^{(t)}}
\end{equation}

This scaling ensures that the effective sample size grows linearly with $T$ while properly weighting each sample's contribution according to its importance in exploring the semantic space.

The modified likelihood for computing $p(\mathcal{D}|\mathbf{p}, K)$ in Equation (13) becomes:
\begin{equation}
    p(\mathcal{D}|\mathbf{p}, K) = \frac{\Gamma(N)}{\prod_{j=1}^K \Gamma({n_j}^{(T)})} \prod_{j=1}^K p_j^{{n_j}^{(T)}}
\end{equation}

\begin{table*}[t]
\centering
\renewcommand{\arraystretch}{1.1}
\setlength{\tabcolsep}{6pt}
\begin{tabular}{L{2.5cm}L{2cm}c@{\hspace{12pt}}NNNN@{\hspace{12pt}}NNNN}
\toprule[1.3pt]
\multirow{2}{*}[-0.8ex]{LLM} & \multirow{2}{*}[-0.8ex]{Dataset} & \multirow{2}{*}[-0.8ex]{P(True)} & \multicolumn{4}{c}{N=2} & \multicolumn{4}{c}{N=5} \\
\cmidrule(lr){4-7} \cmidrule(lr){8-11}
 & & & SAR & SE & SE$_{\text{SDLG}}$ & \textbf{OURS} & SAR & SE & SE$_{\text{SDLG}}$ & \textbf{OURS} \\
\midrule
\multirow{4}{*}{Llama-2-7B} 
 & CoQA       & .468 & .591 & .609 & \underline{.618} & \textbf{.695} & .627 & .683 & \underline{.688} & \textbf{.748} \\
 & TriviaQA   & .488 & .708 & .710 & \underline{.724} & \textbf{.835} & .741 & .795 & \underline{.827} & \textbf{.897} \\
 & TruthfulQA & .509 & .598 & .621 & \underline{.635} & \textbf{.732} & .632 & .713 & \underline{.724} & \textbf{.795} \\
 & SimpleQA   & .521 & .721 & .796 & \underline{.891} & \textbf{.895} & .768 & .930 & \underline{.956} & \textbf{.959} \\
\midrule
\multirow{4}{*}{Llama-3.1-8B} 
 & CoQA       & .568 & .655 & \underline{.699} & .648 & \textbf{.738} & .648 & .756 & \underline{.769} & \textbf{.799} \\
 & TriviaQA   & .714 & .728 & \underline{.759} & .741 & \textbf{.855} & .563 & \underline{.866} & .850 & \textbf{.913} \\
 & TruthfulQA & .633 & \underline{.661} & .642 & .655 & \textbf{.753} & .603 & .741 & \underline{.750} & \textbf{.818} \\
 & SimpleQA   & .655 & .652 & .825 & \underline{.830} & \textbf{.913} & .614 & .930 & \underline{.934} & \textbf{.942} \\
\midrule
\multirow{4}{*}{Mistral-Small-24B} 
 & CoQA       & .618 & .643 & .658 & \underline{.669} & \textbf{.762} & .660 & .758 & \underline{.765} & \textbf{.825} \\
 & TriviaQA   & .624 & .641 & .689 & \underline{.703} & \textbf{.872} & .778 & .790 & \underline{.817} & \textbf{.928} \\
 & TruthfulQA & .597 & .622 & .661 & \underline{.673} & \textbf{.773} & .663 & .765 & \underline{.772} & \textbf{.839} \\
 & SimpleQA   & .668 & .645 & .725 & \underline{.737} & \textbf{.772} & .681 & .868 & \textbf{.873} & \underline{.871} \\
\bottomrule[1.3pt]
\end{tabular}
\caption{AUROC for hallucination detection on open-form QA datasets across three representative LLMs. N denotes the sampling budget, representing the average number of response samples generated per query for uncertainty estimation. The best results are in \textbf{bold} and the second best is marked with \underline{underline}.}
\label{tab:results}
\end{table*}

\section{Experimental Setup}

\paragraph{Datasets and models} We evaluate on four free-form QA datasets: CoQA \cite{tacl2019coqa}, TriviaQA \cite{acl2017triviaqa}, TruthfulQA\cite{acl2022truthfulqa}, and SimpleQA \cite{corr2024simpleqa}, covering both open-book and closed-book scenarios. All experiments use zero-shot settings. We test three LLMs: Llama-2-7B \cite{corr2023llama2}, Llama-3.1-8B \cite{corr2024llama3}, and Mistral-Small-24B \cite{corr2025mistral}, ensuring generalization across different architectures and scales.

\paragraph{Evaluation} We measure hallucination detection performance using AUROC \cite{pr1997auroc}, treating estimator outputs as binary classification scores. Following \cite{corr2024llmasajudge}, we use GPT-4.1 to judge response correctness with a Pass-All@3 method: sampling three responses per question and marking as hallucination if any response contradicts ground truth. For efficiency comparison, we evaluate AUROC at sampling budgets N=1 to 10. As our method uses adaptive sampling, we calibrate variance thresholds to match baseline sample counts.

\paragraph{Baselines} We compare against four methods: \textbf{P(True)} \cite{corr2022ptrue} uses LLM self-assessment; \textbf{SAR} \cite{acl2024sar} aggregates token-level prediction entropy; \textbf{SE} \cite{nature2024semanticentropy} clusters semantic equivalents and computes entropy; \textbf{SE$_{\text{SDLG}}$} \cite{iclr2025sdlg} enhances SE with targeted perturbations for diverse outputs.

\paragraph{Implementation.} We conduct our experiments on a single A800 80GB GPU. Semantic clustering uses DeBERTa-v3-large \cite{iclr2023deberta} for NLI-based equivalence detection. All generation uses temperature 1.0. $N_0$ is set to 1 to reduce the initialization overhead, and top-$k$ is set to 3. Adaptive sampling variance thresholds are calibrated to achieve average sample counts comparable to baselines.

\section{Results and Analyses}

\subsection{Main Results}

Since the computational overhead beyond LLM sampling costs is negligible for all methods (detailed analysis in Appendix C), we set equivalent sampling budgets across different approaches to ensure fair comparison. Table 1 presents the performance comparison across multiple models and datasets. \textbf{First}, our method achieves the highest AUROC in 23 out of 24 settings, with up to 16.9\% improvement over the strongest baseline SE$_{\text{SDLG}}$ (TriviaQA, Mistral-Small-24B, N=2). \textbf{Second}, the advantage is most pronounced in low-sample regimes: 12.6\% average improvement at N=2 versus 6.3\% at N=5, confirming that our adaptive sampling efficiently explores semantic space under computational constraints. \textbf{Third}, consistent improvements across different model architectures and QA domains demonstrate that our method captures fundamental semantic uncertainty properties rather than dataset-specific patterns, ensuring robust practical deployment.

\subsection{Ablation Studies}

We conduct ablation studies on TriviaQA using Llama-3.1-8B to analyze each component's contribution (Table 2).

\paragraph{Prior Estimation} Replacing adaptive prior with fixed $K=K_{obs}+1$ causes 4.3\% (N=2) and 3.5\% (N=5) performance drops, confirming that weighted perplexity effectively captures prompt-specific semantic diversity.

\paragraph{Adaptive Sampling} Fixed sampling without Bayesian framework degrades performance by 9.6\% (N=2) and 4.7\% (N=5). Adding Bayesian framework to fixed sampling partially recovers performance but still underperforms by 5.7\% and 2.8\%. This shows: (1) Bayesian uncertainty quantification benefits all sampling strategies, and (2) variance-based adaptive sampling is crucial for efficiency, especially in low-sample regimes.

\paragraph{Exploration Strategy} Removing guided exploration decreases performance by 3.2\% (N=2) and 2.1\% (N=5), with larger impact at smaller budgets, confirming that importance sampling explores the semantic space and accelerates semantic discovery under constrained resources.

\begin{table}[t]
\centering
\renewcommand{\arraystretch}{1.0}
\setlength{\tabcolsep}{8pt}
\begin{tabular}{lcc}
\toprule
Method & N=2 & N=5 \\
\midrule
\textbf{Full Model} & \textbf{.855} & \textbf{.913} \\
\midrule
\multicolumn{3}{l}{\textit{Prior Estimation}} \\
\quad w/o adaptive prior ($K=K_{obs}+1$) & .812 & .878 \\
\midrule
\multicolumn{3}{l}{\textit{Adaptive Sampling}} \\
\quad Fixed sampling w/o Bayesian & .759 & .866 \\
\quad Fixed sampling w/ Bayesian & .798 & .885 \\
\midrule
\multicolumn{3}{l}{\textit{Exploration Strategy}} \\
\quad w/o guided exploration & .823 & .892 \\
\bottomrule
\end{tabular}
\caption{Ablation study on TriviaQA using Llama-3.1-8B, showing AUROC performance when removing key components of our method.}
\label{tab:ablation}
\end{table}

\begin{table}[t]
\centering
\small
\renewcommand{\arraystretch}{1.1}
\setlength{\tabcolsep}{3pt}
\begin{tabular}{lcccc}
\toprule
Model & CoQA & TriviaQA & TruthfulQA & SimpleQA \\
\midrule
Llama-2-7B & 27.4\% & 37.5\% & 58.8\% & 98.5\% \\
Llama-3.1-8B & 15.9\% & 34.7\% & 52.2\% & 76.2\% \\
Mistral-Small-24B & 13.7\% & 21.2\% & 46.9\% & 71.5\% \\
\bottomrule
\end{tabular}
\caption{Hallucination rates across different models and QA datasets using the Pass-All@3 evaluation method.}
\label{tab:hallucination}
\end{table}

\begin{figure}[ht]
\centering
\begin{subfigure}[b]{0.49\columnwidth}
    \centering
    \includegraphics[width=\textwidth]{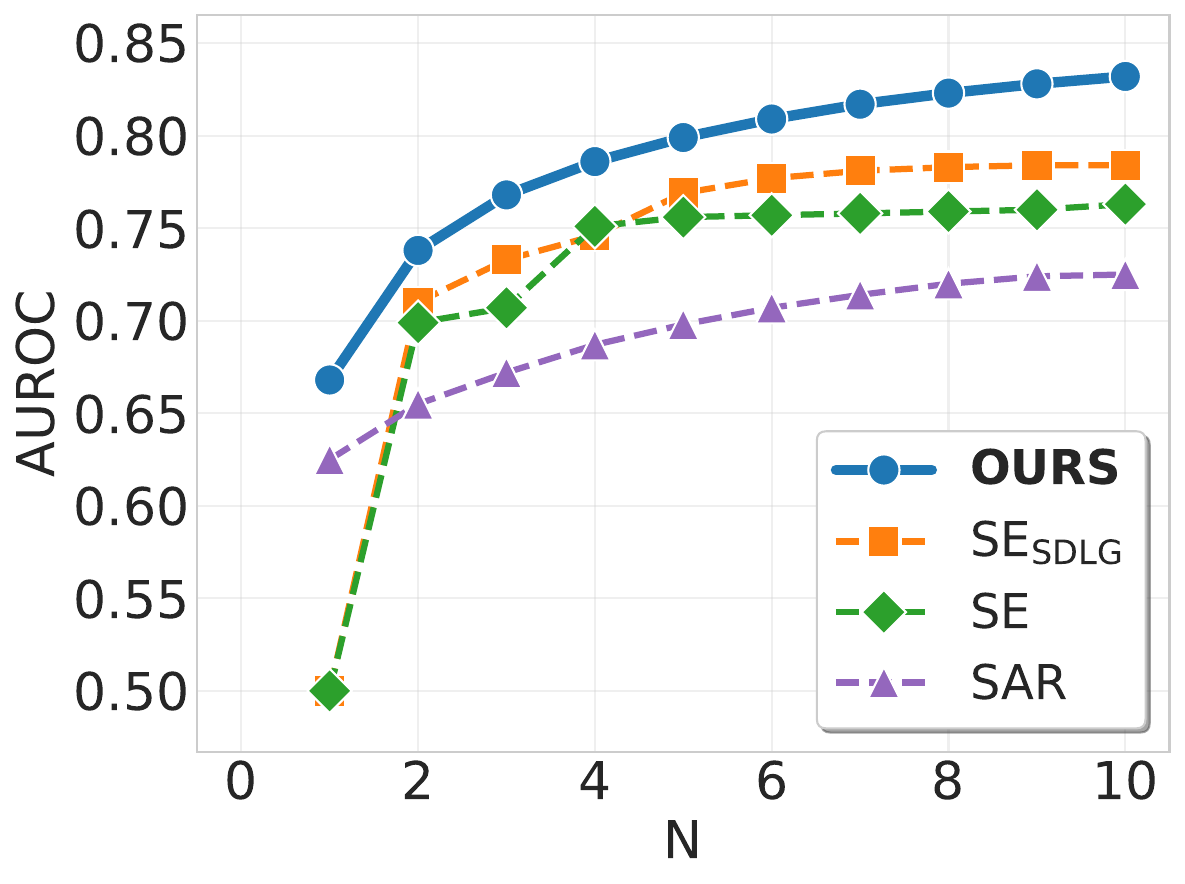}
    \caption{CoQA}
\end{subfigure}
\hfill
\begin{subfigure}[b]{0.49\columnwidth}
    \centering
    \includegraphics[width=\textwidth]{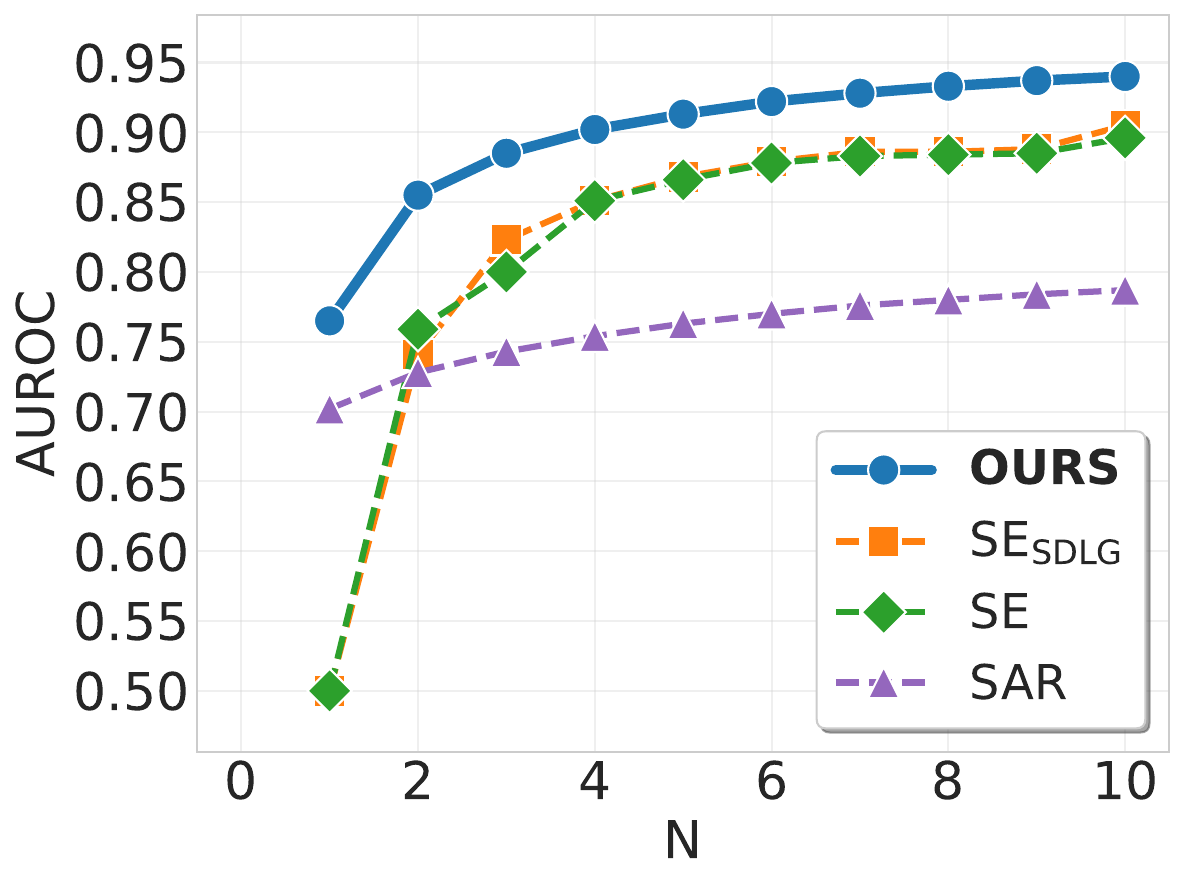}
    \caption{TriviaQA}
\end{subfigure}

\begin{subfigure}[b]{0.49\columnwidth}
    \centering
    \includegraphics[width=\textwidth]{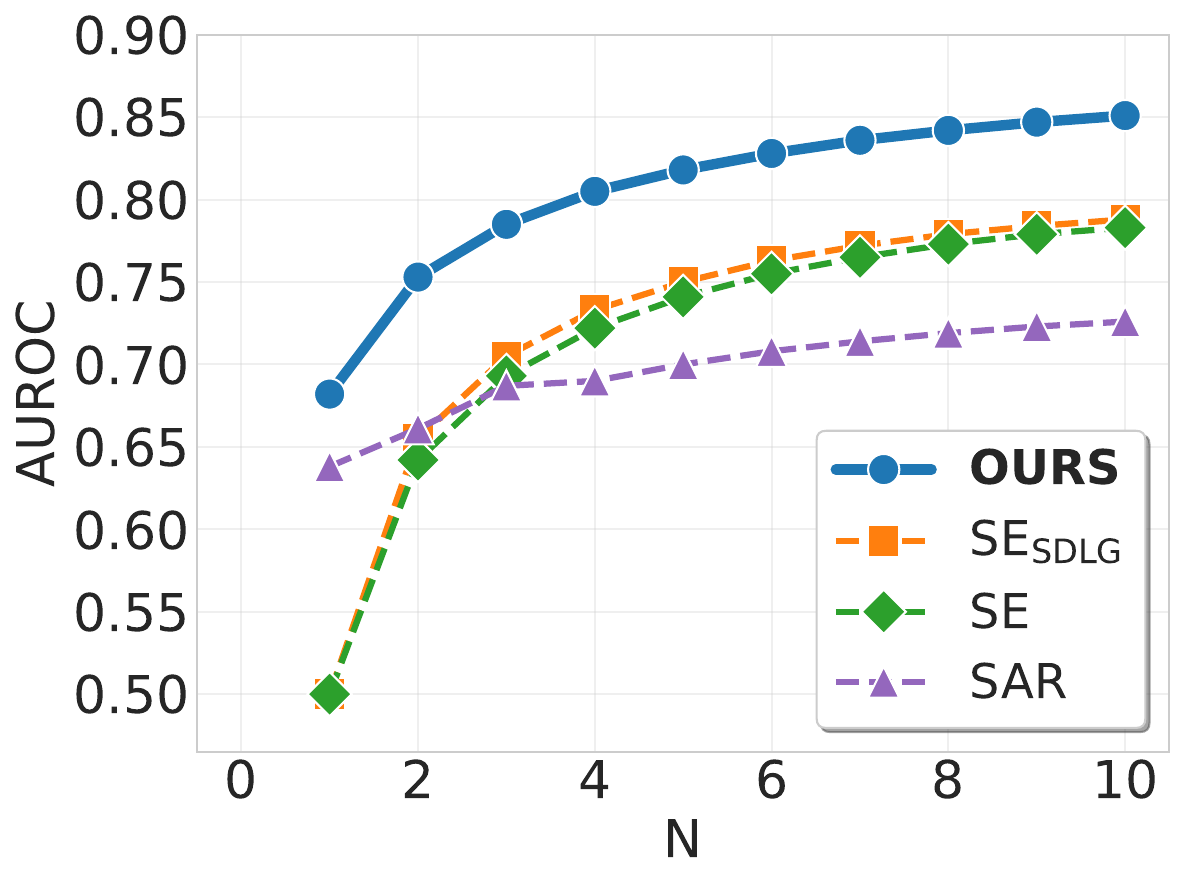}
    \caption{TruthfulQA}
\end{subfigure}
\hfill
\begin{subfigure}[b]{0.49\columnwidth}
    \centering
    \includegraphics[width=\textwidth]{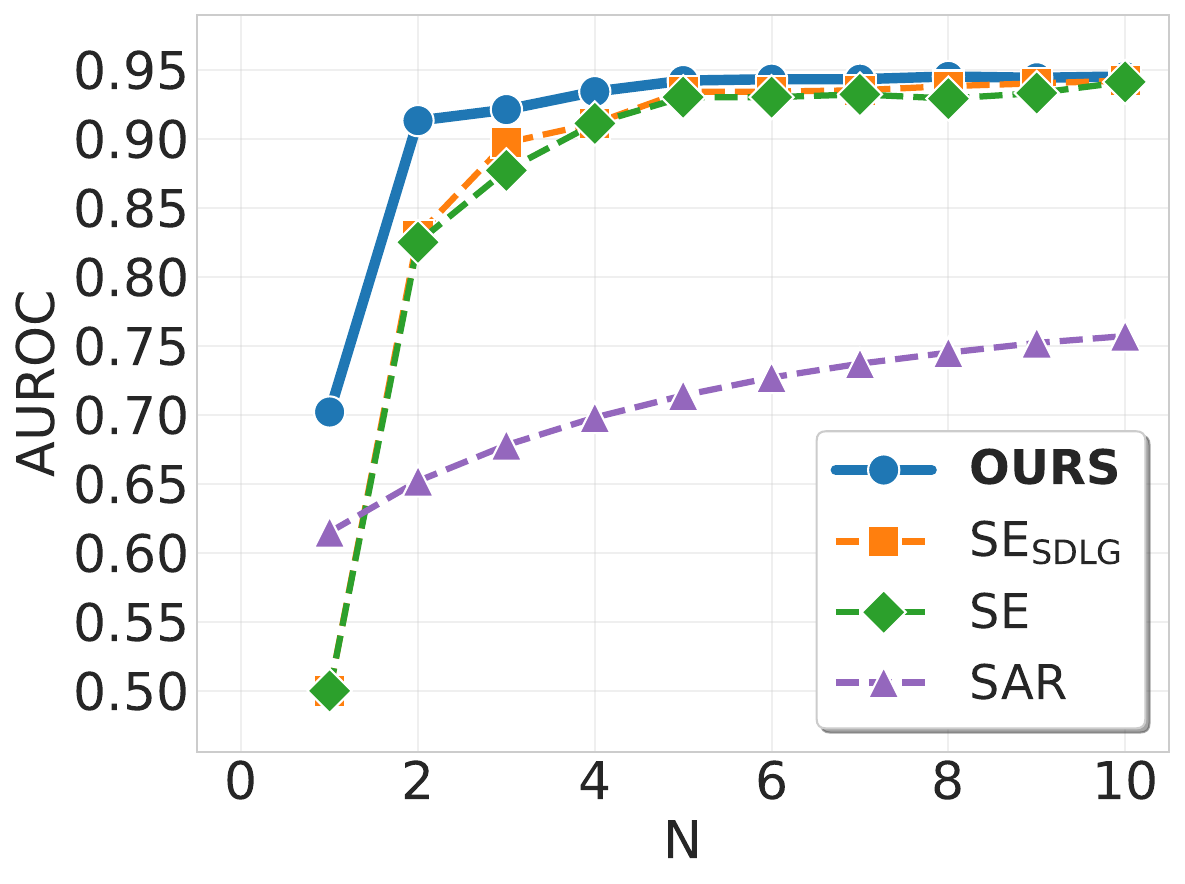}
    \caption{SimpleQA}
\end{subfigure}

\caption{AUROC performance comparison of hallucination detection methods on Llama-3.1-8B across varying sampling budgets N.}
\end{figure}

\begin{figure}[t]
\centering
\includegraphics[width=.95\columnwidth]{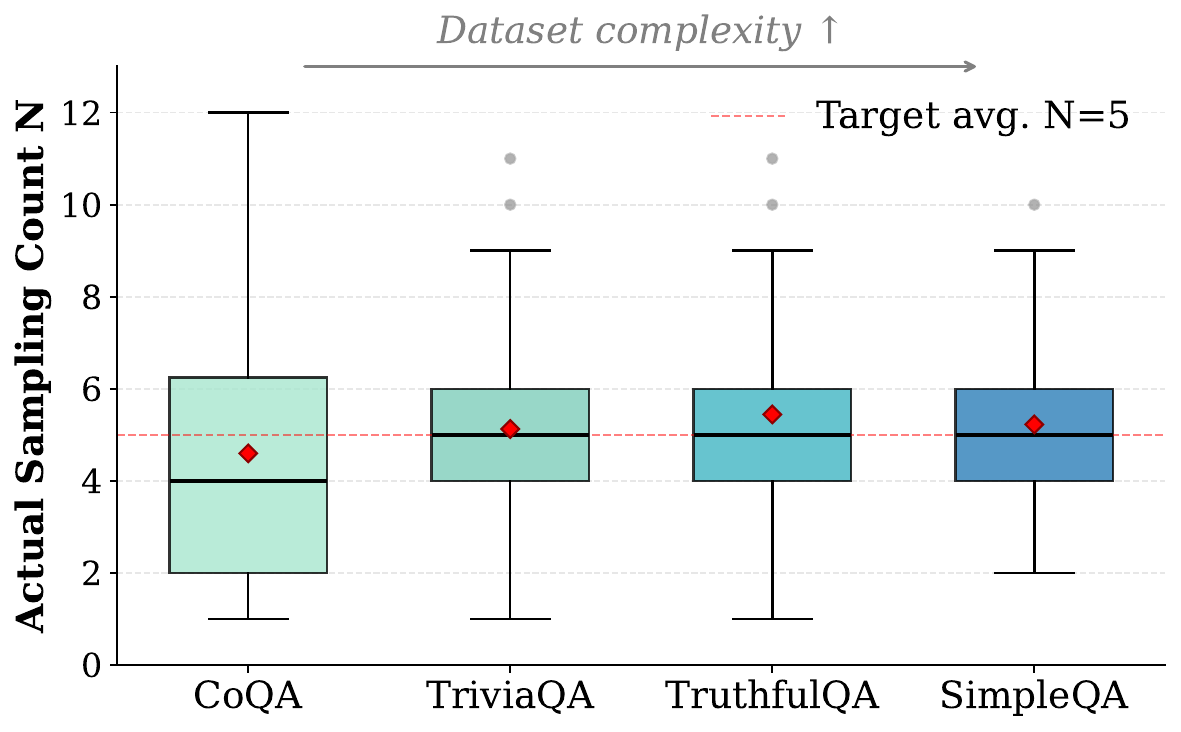}
\caption{Distribution of actual sampling counts under a fixed average budget of N=5 across four QA datasets using Llama-3.1-8B.}
\end{figure}

\subsection{Further Analyses}

\paragraph{Bayesian Framework as a Superior Estimator} Our method maintains advantages even at higher sampling budgets (Figure 2) due to hierarchical uncertainty modeling: semantic space cardinality via $p(K|\mathcal{D})$ and probability distributions via truncated Dirichlet posteriors. Unlike traditional methods assuming fixed semantic categories, we account for unobserved meanings through posterior inference. Generation probability constraints (Equation 6) further tighten the feasible space, improving accuracy even when extensive sampling would reveal most variations.

\paragraph{Adaptive Resource Allocation} Figure 3 shows our adaptive strategy allocating resources by query complexity. Simple datasets (CoQA) exhibit high variance—many queries need only 1-2 samples due to rapid convergence. Complex datasets (SimpleQA) concentrate near the budget limit, indicating persistent uncertainty. This aligns with intuition: high-probability, consistent responses trigger early termination through decreased posterior variance, ensuring efficiency without sacrificing accuracy for real-world queries.

\section{Conclusions}

In this paper, we propose an adaptive Bayesian estimation framework for semantic entropy that addresses the computational inefficiency of existing hallucination detection methods. Our hierarchical approach models semantic distributions through a Dirichlet prior, while guided exploration with importance sampling discovers diverse interpretations. Variance-based adaptive sampling dynamically allocates resources according to query complexity. Experiments demonstrate consistent superiority across multiple models and datasets, with significantly fewer samples required, particularly in low-budget scenarios. The framework's efficient resource allocation makes it practical for real-world deployment. Future directions include extensions to multimodal settings and broader uncertainty quantification tasks.

\section*{Acknowledgments}
This work was supported by the National Natural Science Foundation of China under Grant 62306331 and CAAI Youth Talent Lifting Project under Grant CAAI2023-2025QNRC001.

\bibliography{aaai2026}

\end{document}